\documentclass{article}
\usepackage{spconf,amsmath,graphicx,multirow}
\usepackage{amssymb}
\usepackage{tikz,pgfplots}
\usepackage{bm}
\usetikzlibrary{spy}

\newcommand{\tr}{{\rm tr}}

\usepackage[utf8]{inputenc} %
\usepackage[T1]{fontenc}    %
\usepackage{hyperref}       %
\usepackage{url}            %
\usepackage{booktabs}       %
\usepackage{amsfonts}       %
\usepackage{nicefrac}       %
\usepackage{microtype}      %
\usepackage{color}
\usepackage{enumitem,kantlipsum}
\usepackage{caption}
\usepackage{subcaption}
\usepackage{lipsum} %
\usepackage{multicol} %

\newcommand{\haf}[1]{\textcolor{black}{#1}}
\newlength{\Oldarrayrulewidth}
\newcommand{\Cline}[2]{%
  \noalign{\global\setlength{\Oldarrayrulewidth}{\arrayrulewidth}}%
  \noalign{\global\setlength{\arrayrulewidth}{#1}}\cline{#2}%
  \noalign{\global\setlength{\arrayrulewidth}{\Oldarrayrulewidth}}}

\title{Deep neural network Models trained with a fixed Random classifier transfer better across domains}

\name{Hafiz Tiomoko Ali{$^{1}$,} Umberto Michieli{$^{1}$,} Ji Joong Moon{$^{2}$,}  Daehyun Kim{$^{2}$,} Mete Ozay{$^{1}$}}
\address{$^{1}$\,Samsung Research UK, $^{2}$\,Samsung Research Korea}
\begin{document}
\maketitle
\begin{abstract}
The recently discovered Neural collapse (NC) phenomenon states that the last-layer weights of Deep Neural Networks (DNN), converge to the so-called \emph{Equiangular Tight Frame (ETF)} simplex, at the terminal phase of their training. This ETF geometry is equivalent to vanishing within-class variability of the last layer activations. Inspired by NC properties, we explore in this paper the transferability of DNN models trained with their last layer weight fixed according to ETF. This enforces class separation by eliminating class covariance information, effectively providing implicit regularization. We show that DNN models trained with such a fixed classifier significantly improve transfer performance, particularly on out-of-domain datasets. 
On a broad range of fine-grained image classification datasets, our approach outperforms $i)$ baseline methods that do not perform any covariance regularization (up to $22\%$), as well as $ii)$ methods that explicitly whiten covariance of activations throughout training (up to $19\%$). Our findings suggest that DNNs trained with fixed ETF classifiers offer a powerful mechanism for improving transfer learning across domains.
\end{abstract}
\section{Introduction}
\label{sec:intro}

Deep neural networks (DNNs) have revolutionized machine learning and pattern recognition tasks by providing highly expressive feature extractors. Recent trends in training DNN models involve the use of fixed classifiers\cite{lu2020neural, yang2022classifier, yang2023neural} instead of traditional trainable linear layers on top of features extracted from the last layer of the DNNs. This approach is inspired by the Neural Collapse (NC) phenomenon~\cite{papyan2020prevalence, gao2023study}, which occurs during the late stages of model training. NC reveals that features obtained from the last layer of the network converge to class means with their within-class variability collapsing to zero and forming the so-called Equiangular Tight Frame (ETF) geometry; the last-layer linear classifiers are perfectly matched with their class activation means, resulting in classifiers with ETF geometry. 

\begin{figure}
    \centering
    \includegraphics[trim=0cm 15cm 19cm 0cm, clip, width=\linewidth]{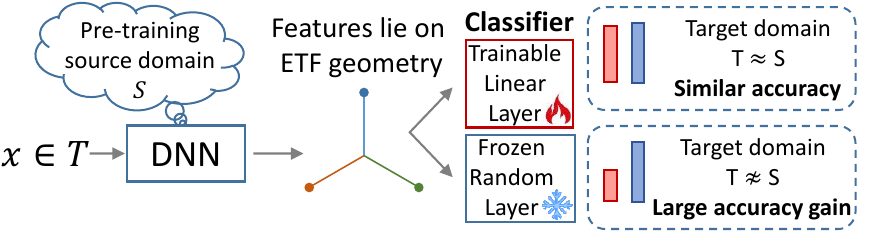}
    \caption{DNN models pretrained on a source domain $S$ with a classifier following the ETF geometry implicitly minimize class features variability. This translates into increased transferability to out-of-distribution target domains $T$.}
    \label{fig:acquisition_sequences}
\end{figure}

In this paper, we explore the transferability of training DNN models with fixed classifiers following an ETF geometry. The ETF structure implicitly enforces negligible within-class variability of features, thus enhancing cluster separability which in turn, significantly improves transfer performance. Our experiments focus on evaluating these models on various fine-grained image classification datasets, and especially on out-of-domain scenarios. Our contributions can be summarized as follows:

\begin{enumerate}
\item Through the use of existing results in Random Matrix Theory (RMT), we establish that the utilization of a linear random projector adheres to ETF classifiers properties, leading to a feature kernel with minimal class covariance.
\item We then empirically show that DNNs trained with fixed ETF classifiers lead to enhanced transfer performance on a range of fine-grained image classification datasets, especially on those with important domain shifts with respect to the source dataset (see Figure~\ref{fig:acquisition_sequences}).
\item We compare and show superior performance of our approach with baseline methods that train the linear classifier from scratch (\emph{no covariance minimization}), as well as methods such as SW (Switchable Whitening)~\cite{pan2019switchable}, that explicitly perform features covariance whitening during training.
\end{enumerate}
{\bf Notations.} Vectors (matrices) are denoted by lowercase (uppercase) boldface letters.
$\left\{{\bf v}_a\right\}_{a=1}^n$
is the column vector ${\bf v}$ with (scalar or vector) entries ${\bf v}_a$ and $\left\{{\bf V}_{ab}\right\}_{a,b=1}^n$ is the matrix
${\bf V}$ with (scalar or matrix) entries ${\bf V}_{ab}$. The vector ${\bf 1} \in \mathbb{R}^n$ stands for the column vector
filled with ones. The vector ${\bf j}_a$ is the canonical vector of class $\mathcal{C}_a$
defined by $({\bf j}_a)_i = \delta_{{\bf x}_i \in \mathcal{C}_a}$ and ${\bf J} = [{\bf j}_1,\ldots, {\bf j}_K] \in \{0,1\}^{N\times K}$.

\section{Related work}

The NC phenomenon, as illuminated by recent research~\cite{papyan2020prevalence, han2021neural, zhu2021geometric}, sheds light on the evolving behavior of features in DNN models during the late stages of training. It posits that more expressive models possess the capability to selectively discard class-covariance. Building upon the insights from NC, recent contributions~\cite{yang2022classifier, yang2023neural} have harnessed the properties of NC by utilizing fixed random classifiers to train backbone  models. These models are designed to generate features that align with the geometric structure of the fixed random classifier. Notably, these studies have demonstrated improved performance, particularly when dealing with training data exhibiting inherent class imbalance. In this paper, we go further by showcasing that models equipped with this mechanism also exhibit significant enhanced transferability, especially when faced with out-of-distribution target data. This remarkable improvement in transferability can be attributed to the trivial separation of features per class, achieved by eliminating class covariances during training.

In contrast to numerous methods that explicitly whiten features covariances (enforce covariance matrix to be close to the identity matrix) during training using dedicated loss functions~\cite{pan2019switchable}, our approach focuses on enforcing negligeable within-class variability, throughout training. Whitening operations discard the variability between all features within a batch, while the use of fixed ETF classifier adhere to the properties of NC by implicitly discarding \emph{within-class features variability}.

Our work bridges the application of fixed ETF classifiers with the use of Random Projection (RP) classifiers. Random Matrix Theory has proven invaluable for analyzing the behavior of various machine learning methods, in particular RP~\cite{rahimi2007random}. In a related study~\cite{ali2021random}, kernels obtained from random projectors with generic activation functions were analyzed. However, this study operates under the Gaussian Mixture Model (GMM) assumption, aiming to achieve "non-trivial" separation based on class means and class covariances. We establish that the utilization of a linear random projector adheres to ETF classifiers properties, leading to a feature kernel with minimal class covariance. This observation rationalizes the use of fixed random classifiers adhering to the ETF geometry, which can essentially be seen as a random projector with a linear activation function.

In summary, our study extends the potential of fixed ETF classifiers by showcasing their effectiveness in cross-domain transfer tasks. This research offers a perspective of the role played by fixed ETF classifiers in feature transformation and transfer learning, particularly when confronted with diverse data distributions.

\section{Transfer Learning with Fixed ETF Classifiers}
\label{sec:pagestyle}
\subsection{Linear random features exhibit minimal class covariance and thus achieve NC}
In this section, we aim to establish the equivalence between the Neural Collapse (NC) phenomenon and the use of linear random features for enhancing classification robustness and generalization. Our goal is to elucidate the design of a random projector that transforms features learned by DNN models into a space where they are linearly separable.

To this end, we consider a classification task with $K$ classes $\{ \mathcal{C}_k\}_{k=1}^K$, where $n$ training features $\{ {\bf x}_i \in \mathbb{R}^p \}_{i=1}^n$  belong to one of the $K$ classes. Our objective is to demonstrate the design of a random projector that transforms these features into a space where they are linearly separable. These features are extracted from a pretrained DNN model, allowing us to model them using a Gaussian Mixture Model (GMM) as shown in previous works~\cite{seddik2020random}. This means that $\{ {\bf x}_i \}_{i=1}^n$ are independent vectors belonging to $\{ \mathcal{C}_k\}_{k=1}^K$, such that ${\bf x}_i \in \mathcal{C}_a$ can be modeled as
\begin{equation*}
\label{eq:xi}
{\bf x}_i = {\bm \mu}_a + {\bf C}_a^{\frac12}{\bf z}_i
\end{equation*}
where ${\bf z}_i \in \mathbb{R}^p$ is uniformly distributed at random on the $\sqrt{p}$-radius sphere $\mathcal{S}^{p-1}$ of $\mathbb{R}^p$, and ${\bm \mu}_a \in \mathbb{R}^p$, ${\bf C}_a \in \mathbb{R}^{p\times p}$ are the deterministic class means and class covariances of ${\bf x}_i$.

The Neural Collapse (NC) phenomenon~\cite{papyan2020prevalence} occurs when several conditions are met for features extracted using a DNN model $f$. These conditions are: 
$i)$ The within-class variability of features obtained before the last layer of $f$ converges to zero.
$ii)$ The features extracted from samples of each class converge to their class means, forming an ETF.
$iii)$ The normalized weight vector corresponding to each class features extracted at the last linear layer of $f$ converges to the corresponding class mean of the features.
Now, let ${\bf X}=[{\bf x}_i]_{i=1}^n \in \mathbb{R}^{p \times n}$, ${\bf M}=[{\bm \mu}_i]_{i=1}^K \in \mathbb{R}^{p \times K}$, and ${\bf C}^{\circ} = \sum_{a=1}^K \frac{n_a}{n} {\bf C}_a$. We can associate the NC properties as follows:
\begin{enumerate}
    \item Property $i)$ is equivalent to ${\bf C}^{\circ} \to  {\bf 0}$.
    \item Property $ii)$ is equivalent to ${\bf X}^T{\bf X} \to {\bf M}^T{\bf M}$.
    \item Property $iii)$ states that the weight matrix of the classifier is ${\bf W} = \sqrt{\frac{K}{K-1}}{\bf PQ}$ where ${\bf P}={\bf I}_K-\frac1K {\bf 1}_K{\bf 1}_K^T$ and ${\bf Q} \in \mathbb{R}^{K \times p}$ is a random orthogonal matrix.
\end{enumerate}

A linear classifier with the weights ${\bf W}$ is called a \textit{fixed ETF classifier}.

Our objective is to establish the equivalence between the properties of NC and the use of linear random features. To this end, we define random features by $\sigma({\bf WX})$ with ${\bf W}$ and ${\bf X}$ defined earlier, and $\sigma(\cdot)$ a generic activation function. Our goal is to prove that a linear activation $\sigma(t)=t$ achieves NC.

We make use of existing RMT results in this context. While various RMT algorithms~\cite{ali2018random, liao2019inner, couillet2016kernel} have been developed to discriminate both class means and class covariances of features ${\bf x}_i$, it turns out that for the final stages of training, discriminating only the class means is sufficient to improve robustness and generalization of classifiers.

Since the properties of NC are summarized in terms of features covariance, we define the kernel (random feature covariance) by
$
    {\bf K} = \frac1p \mathbb{E}_{{\bf W}}\sigma({\bf WX})^T\sigma({\bf WX}), 
$
where the expectation is over the set of orthogonal random matrices.
Under specific growth rates (with respect to the dimension of features) assumptions on class mean and class covariance functionals,  ~\cite{ali2021random} provides a random equivalent of ${\bf K}$ as a function of class means and covariances, for a broad class of activation functions that satisfy boundedness of their first and second derivatives.  Theorem~$1$ in~\cite{ali2021random} states that the kernel corresponding to random features can be decomposed into two terms:
\begin{enumerate}[noitemsep,nolistsep,leftmargin=*]
    \item $d_1 {\bf X}^T{\bf X}=d_1 \cdot \left({\bf Z}+{\bf M}\frac{{\bf J}^{\sf T}}{\sqrt{p}}\right)^{\sf T}\left({\bf Z}+{\bf M}\frac{{\bf J}^{\sf T}}{\sqrt{p}}\right)$ which is essentially the kernel of the features when no random projection is applied. This means that the kernel term $d_1 {\bf X}^T{\bf X}$ retains the inherent structure of the features ${\bf X}$.
    \item $d_2 \cdot \left({\bm \phi}+{\bf VAV}^{\sf T}\right)$ only containing information about the class covariances ($\tr {\bf C}_a^\circ, \tr {\bf C}_a^\circ{\bf C}_b^\circ$). This means that the kernel term responsible for class covariances vanishes.
\end{enumerate}
\hspace{10pt}

According to Property $i)$ of NC, ${\bf C}_a^\circ \to 0$ (as ${\bf C}^{\circ}= \sum_{a=1}^K \frac{n_a}{n} {\bf C}_a \to {\bf 0}$). Consequently, we want $d_2=0$ to achieve Property $i).$ Regarding Property $2,$ as long as the features are extracted from a stronger feature extractor, $d_1 {\bf X}^T{\bf X}$ will be closer to ${\bf M}^T{\bf M}$. In other words, a stronger feature extractor will have a lower energy for the matrix ${\bf Z}$ compared to the signal matrix ${\bf M}$. For a linear activation function $\sigma(t)=t,$ $d_1=1$ and $d_2=0$ implying that training a strong feature extractor along with a fixed linear random projector, results in maximal class separation. Note that ${\bf Z}$ contains random Gaussian columns with zero mean and covariance ${\bf C}_a$. The use of a linear random classifier to train the feature extractor forces ${\bf C}_a$ to be negligible.

\begin{figure}
\centering
 \begin{tikzpicture}[font=\footnotesize][scale=0.5]
\pgfplotsset{every major grid/.style={style=densely dashed}}
    		\begin{axis}
        [width=0.4\textwidth,height=0.4\textwidth, grid=major,xlabel={$a_2$},ylabel={MSE}, xmin=-1.1,xmax=1.1,ymin=0.37,ymax=1.38,  xtick={-1, 0, 1},
    xticklabels={$-1$, $0$, $1$}]

 \addplot[mark size=4pt, blue, dashed, mark=star,thick]coordinates{            
(-1, 1.3680362)(-0.8, 1.24733223)(-0.6, 1.08262322)(-0.4, 0.8614634)(-0.2, 0.56958427)(0, 0.37451134)(0.2, 0.56958427)(0.4, 0.8614634)(0.6, 1.08262322)(0.8, 1.24733223)(1, 1.3680362)
};
    		\end{axis}
    		\end{tikzpicture}   
    \caption{
    Test MSE. Random feature regression with activation function $\sigma(t)=a_2t^2 + t$. Features extracted from the STL dataset using RN50 model pretrained on Imagenet. }
\label{fig:mse-ridge-reg}
\end{figure}
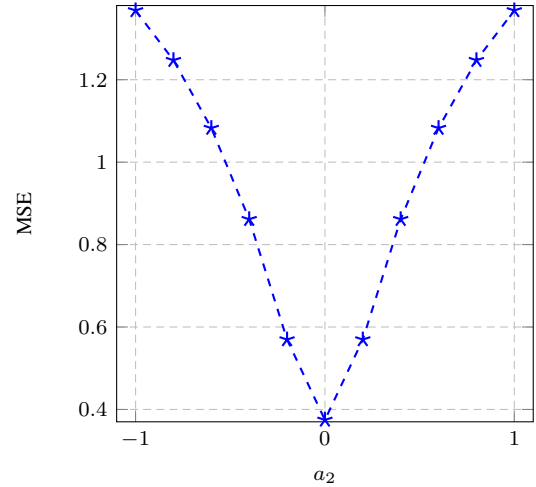

 \begin{figure}[t!] %
    \centering
    \includegraphics[width=\linewidth]{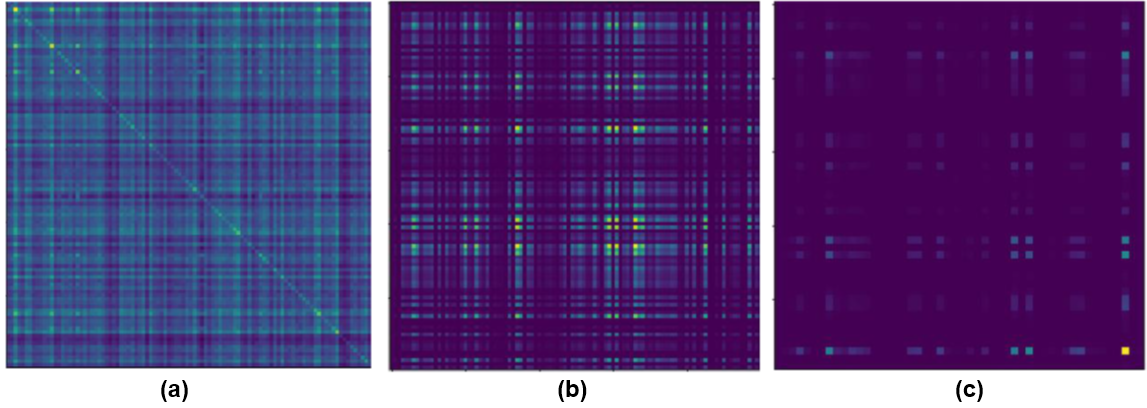}
    \caption{Heatmap of the covariance matrices ${\bf C}^\circ$ of features obtained after transfer learning on the Flowers dataset using the RN101 model pretrained on the ImageNet dataset. $a)$ Trainable model with $\frac1p \tr {\bf C}^\circ = 0.16$,  $b)$ {SW} with $\frac1p \tr {\bf C}^\circ = 0.04$ and $c)$ {Fixed} model with $\frac1p \tr {\bf C}^\circ = 0.01$.}
    \label{fig:heatmaps}
\end{figure}

\subsection{Practical implications for transfer learning}
We first illustrate the observations in the previous section in a transfer regression setup. The goal is to show that, in that scenario, linear random features (i.e., $\sigma(t)=t$) transfer better than non-linear random features (i.e., $\sigma(t) \ne t$). We train the ResNet50 (RN50) model on the source domain dataset $S$ (ImageNet)~\cite{imagenet_cvpr09}. We then extract features from images of the target domain $T$ (STL dataset)~\cite{coates2011analysis}. We then apply a regression model on $\sigma({\bf WX})$ with polynomial activation function $\sigma(t)=a_2 t^2 + t$ to the target features. Figure~\ref{fig:mse-ridge-reg} shows the test Mean Squared Error (MSE) as a function of $a_2$~\footnote{Note that for a polynomial activation function $\sigma(t)=a_2t^2 + a_1t,$ $d_1=a_1^2$ and $d_2=2a_2^2$.}. We can see from Figure~\ref{fig:mse-ridge-reg} that the linear activation is optimal when using the \emph{expressive} feature extractor RN50.

Next, we consider transfer learning experiments on DNN models. Our transfer learning framework follows the illustration in Figure~\ref{fig:acquisition_sequences}. A DNN model was pretrained on a source dataset $S$ using two strategies: $i)$ classical pretraining where a linear classifier is trained along the DNN's backbone; $ii)$ only the backbone is trained with the classifier fixed as a linear random layer following ETF geometry. We then finetune the pretrained model on a target dataset $T.$ 
We conducted a series of experiments to evaluate the transferability of DNNs trained with fixed ETF classifiers. Our main objective is to demonstrate the superior performance of these models when transferring to new domains, especially when dealing with out-of-distribution datasets.

We employed two DNN architectures: ResNet50 and ResNet101, and pretrained them on the popular ImageNet dataset. Subsequently, we performed transfer experiments on the following target datasets: CIFAR100~\cite{krizhevsky2009learning}, CIFAR10~\cite{krizhevsky2009learning}, STL10~\cite{coates2011analysis}, Oxford Pets~\cite{parkhi2012cats}, Oxford Flowers~\cite{nilsback2006visual}, DTD~\cite{cimpoi14describing},  SVHN~\cite{netzer2011reading}.%

We pretrained both ResNet50 and ResNet101 \cite{he2016deep} backbones using three different strategies:
\begin{enumerate}
    \item \textit{Trainable Model}: Training with a trainable linear classifier.
    \item \textit{SW}: Training with the Switchable Whitening~\cite{pan2019switchable} method that explicitly whiten class covariances during pretraining.
    \item \textit{Fixed Model (Ours)}: Training with a fixed ETF classifier, enforcing class separation through implicit minimization of class covariances.
\end{enumerate}

RN50 models were trained for $180$ epochs and RN101 were pretrained for $100$ epochs with an initial learning rate of 0.01, a batch size of 128, momentum of 0.9, and weight decay of $2 \times 10^{-4}$. We applied standard data augmentation and normalization for each corresponding dataset. 

For the transfer learning experiments, we fine-tuned each model on the target datasets, starting from the checkpoint of the aforementioned methods. Table~\ref{tab:transfer_results_train_imagenet} summarizes the results of these experiments.

\begin{table}[t!]
\setlength{\tabcolsep}{3.5pt}
\centering
\caption{Analyses for transfer learning using RN50/RN101 models pretrained on the ImageNet \haf{(Source {\bf S})} and finetuned on the transfer datasets \haf{(Target {\bf T})}.}
\label{tab:transfer_results_train_imagenet}
\begin{tabular}{cccccc}
\toprule
& \textbf{Transfer data} & \textbf{Model} & \textbf{RN50} & \textbf{RN101} & \\
\midrule
\multirow{11}{*}{\rotatebox[origin=c]{90}{\textbf{In Domain}}} & \multirow{2}{*}{ImageNet \haf{({\bf S})}} & Trainable Model & $75.25$ & ${\bf 79.20}$\\
& & SW & ${\bf 77.93}$ & $79.13$\\
& & Fixed Model (ours) & $76.72$ & $77.29$ \\\cline{2-5}
& \multirow{3}{*}{CIFAR100 \haf{({\bf T})}} & Trainable Model & $61.59$ & $59.86$\\
& & SW & ${\bf 62.55}$ & $60.30$\\
& & Fixed Model (ours) & $62.33$  & ${\bf 62.38}$\\\cline{2-5}
& \multirow{3}{*}{CIFAR10 \haf{({\bf T})}} & Trainable Model & ${\bf 89.56}$ & $87.93$\\
& & SW & $88.68$ & ${\bf 88.43}$ & \\
& & Fixed Model (ours) & $89.33$ &  $88.38$\\\cline{2-5}

& \multirow{3}{*}{STL10 \haf{({\bf T})}} & Trainable Model & $60.03$  &  $56.32$ \\
& & SW & $63.32$  &  $61.91$ \\
& & Fixed Model (ours) & ${\bf 63.73}$ & ${\bf 65.17}$ & \\\Cline{1pt}{2-5}

\multirow{12}{*}{\rotatebox[origin=c]{90}{\textbf{Out of Domain}}}&  \multirow{3}{*}{SVHN \haf{({\bf T})}} & Trainable model & $95.10$  &  $95.56$\\
& & SW & ${\bf 95.67}$  &  ${\bf 95.67}$\\
& & Fixed Model (ours) & $94.40$  & $95.02$ &\\\cline{2-5}
& \multirow{3}{*}{Pets \haf{({\bf T})}} & Trainable Model & $40.50$   &  $46.47$\\
& & SW & $39.22$   & $45.10$ \\
& & Fixed Model (ours) & ${\bf 48.32}$ &  ${\bf 53.17}$\\\cline{2-5}
& \multirow{3}{*}{Flowers \haf{({\bf T})}} & Trainable model & $38.05$ &  $37.09$ \\
& & SW & $41.56$ & $44.07$  \\
& & Fixed Model (ours) & ${\bf 60.30}$ & ${\bf 57.03}$ \\\cline{2-5}
& \multirow{3}{*}{DTD \haf{({\bf T})}} & Trainable model & $29.09$ &  $28.24$ \\
& & SW & $24.62$   & $30.74$\\
& & Fixed Model (ours) & ${\bf 40.90}$ &  ${\bf 34.68}$\\\bottomrule

\end{tabular}
\end{table}

The results presented in Table~\ref{tab:transfer_results_train_imagenet} highlight the effectiveness of utilizing DNN models trained with fixed ETF classifiers for transfer learning tasks. We have split the results into ``In-domain" and ``Out-Of-Domain" categories to showcase the difference in accuracy of models between both domain-specific and domain-agnostic transfer learning tasks.

{\bf In-Domain Transfer}:
The three models achieve comparable accuracy. When the source dataset $S$ and target dataset $T$ have lower domain shift, minimizing the class covariances during adaptation will have a small effect as the class-covariances of the similar source domain data have already been made negligeable during pretraining.

{\bf Out-Of-Domain Transfer}:
Here, our approach consistently outperforms the baseline methods across all datasets and architectures, with a gain of up to $19\%$ compared to the SW~\cite{pan2019switchable} method. This suggests that the model excels at adapting to different and unrelated data distributions. When the source domain $S$ has already undergone the implicit minimization of class covariances, the fine-tuned model benefits from this property. In out-of-distribution domains, class covariances can significantly vary from the source domain. By implicitly minimizing these covariances during training, the model becomes more robust to domain shifts. It will tend to focus more on relevant features for class separation rather than domain-specific covariances. This is highlighted in Figure~\ref{fig:heatmaps} where the covariance ${\bf C}^\circ{}$ of the extracted features from the finetuned model is closer to ${\bf 0}$ for the fixed ETF model than it is for the baseline models (\emph{Trainable} and \emph{SW}).

\section{Conclusion}
In this paper, we have investigated the concept of utilizing DNN models trained with fixed random classifiers for enhanced transferability in downstream tasks. We have shown that a random projector with a linear activation function, applied to features extracted from DNN models, offers strong transfer performances especially in scenarios involving out-of-distribution domains, on a range of fine-grained image classification tasks. By implicitly minimizing class covariances during pretraining, the models tend to focus on essential features for class separation, reducing dependency on domain-specific variations. This work provides a practical strategy for improving transfer learning on challenging out-of-distribution target domains, opening up avenues for future research in advancing the state-of-the-art on transfer learning with ``foundation models".

\vfill\pagebreak

\bibliographystyle{IEEEbib}
\bibliography{refs}

\end{document}